# ViTaL: An Advanced Framework for Automated Plant Disease Identification in Leaf Images Using Vision Transformers and Linear Projection For Feature Reduction

Abhishek Sebastian [1][0000-0002-3421-1450], Annis Fathima A[1][0000-0002-8769-4222], Pragna R [1][0000-0003-0827-5896], Madhan Kumar S [1][0009-0001-7005-4130], Yaswanth Kannan G [1][0009-0006-1871-4572] and Vinay Murali [2][0009-0003-5724-0515]

[1] School of Electronics Engineering, Vellore Institute of Technology, Kelambakkam, Chennai - 600127

[2] School of Mechanical Engineering, Vellore Institute of Technology, Kelambakkam, Chennai – 600127

`Abhishek.sebastian2020@vitstudent.ac.in`

**Abstract.** Our paper introduces a robust framework for the automated identification of diseases in plant leaf images. The framework incorporates several key stages to enhance disease recognition accuracy. In the pre-processing phase, a thumbnail resizing technique is employed to resize images, minimizing the loss of critical image details while ensuring computational efficiency. Normalization procedures are applied to standardize image data before feature extraction. Feature extraction is facilitated through a novel framework built upon Vision Transformers, a state-of-the-art approach in image analysis. Additionally, alternative versions of the framework with an added layer of linear projection and blockwise linear projections are explored. This comparative analysis allows for the evaluation of the impact of linear projection on feature extraction and overall model performance. To assess the effectiveness of the proposed framework, various Convolutional Neural Network (CNN) architectures are utilized, enabling a comprehensive evaluation of linear projection's influence on key evaluation metrics. The findings demonstrate the efficacy of the proposed framework, with the top-performing model achieving a Hamming loss of 0.054. Furthermore, we propose a novel hardware design specifically tailored for scanning diseased leaves in an omnidirectional fashion. The hardware implementation utilizes a Raspberry Pi Compute Module to address low-memory configurations, ensuring practicality and affordability. This innovative hardware solution enhances the overall feasibility and accessibility of the proposed automated disease identification system. This research contributes to the field of agriculture by offering valuable insights and tools for the early detection and management of plant diseases, potentially leading to improved crop yields and enhanced food security.

# 1    Introduction

Food safety and security are paramount concerns in today's world, where global food supply chains face multiple challenges ranging from wastage to disease identification and novel technological solutions. The magnitude of food wastage, as highlighted in recent studies, is staggering. Approximately one-third of the food produced globally is lost or wasted [1], representing not only a colossal economic loss but also a grave threat to food and nutrition security [5]. It is imperative to address this issue comprehensively to ensure that the world's growing population has access to sufficient and safe food.

One of the critical aspects contributing to food wastage is the presence of plant diseases that affect crops on a large scale. Plant pathogens, ranging from viroids to higher plants, can cause diseases with varying degrees of severity, from mild symptoms to catastrophic losses. Such diseases result in reduced crop yields, exacerbating global food deficits and impacting at least 800 million people who are inadequately nourished [1]. These pathogens pose a complex challenge due to their variability in populations across time, space, and genotype. Additionally, their ability to evolve and overcome plant resistance further complicates the situation.

Addressing plant diseases requires a multifaceted approach encompassing the swift and accurate identification of the causal organism, precise estimation of disease severity and its impact on crop yields, and identification of virulence mechanisms. Conventional plant breeding for resistance, facilitated by marker-assisted selection, plays a crucial role in mitigating these diseases. Furthermore, transgenic modification involving genes conferring resistance has emerged as a promising avenue. [1] discusses these aspects and underscores the importance of acknowledging plant diseases as a substantial threat to global food security.

In addition to plant diseases, the rejection of food exports is a pressing concern affecting food safety and economic stability. Export rejection of food products has led to significant economic losses for many countries, including Indonesia. It is essential to identify the factors contributing to these rejections and estimate the economic losses associated with them [2]. Furthermore, analyzing barriers to sustainable food trade, especially concerning agri-food products, is crucial. In China's case, exports of food to the United States have faced rejections due to concerns about food safety. These rejections have prompted the need for enhanced risk analysis and improved food safety measures [3]. Moreover, understanding the potential for fruit exports and identifying barriers is vital for countries like Sri Lanka [4]. While Sri Lanka has vast potential in the fresh and processed fruit export industry, several obstacles, including low plant health standards, international competition, and quality-related issues, hinder its progress [4].

This research paper focuses on automating defect detection in fruits, vegetables, and food products during packing to prevent unnecessary transportation and reduce the car-bon footprint. The primary emphasis is on plant disease identification using pre-processing tools, including thumbnail resizing and normalization. Feature extraction employs a low-power Vision Transformer (ViT) for subsequent classification with Convolutional Neural Network (CNN) architectures. The investigation includes an auxiliary linear projection for additional feature extraction, with results compared to analyze its impact on prediction outcomes. Furthermore, we present a novel hardware design capable of scanning leaves omnidirectionally to analyze their health.

## 2 Literature Survey

Before the advent of deep learning, traditional computer vision techniques dominated the field of image recognition. These methods relied on handcrafted features, such as Histogram of Oriented Gradients (HOG) and Scale-Invariant Feature Transform (SIFT), combined with machine learning algorithms like Support Vector Machines (SVM) and decision trees [8].

The turning point in image recognition came with the introduction of deep residual learning by He et al. in 2016. They proposed a novel approach to ease the training of deep neural networks by introducing residual connections. Instead of learning unreferenced functions, these networks explicitly reformulated layers as residual functions with reference to layer inputs. This breakthrough allowed the training of networks substantially deeper than before, with depths of up to 152 layers. The residual networks demonstrated their prowess on the ImageNet dataset, achieving a remarkable 3.57% error rate [8].

Moving into the 2020s, the application of deep learning techniques to plant disease identification gained prominence. Thakur et al. conducted a systematic review in 2022, addressing the challenges in this domain [9]. They noted that while deep learning, especially convolutional neural networks (CNNs), had become the standard for image recognition, many methods primarily demonstrated their efficacy on public datasets captured under controlled conditions. One of the contrasting techniques highlighted in their review was the limitation of lightweight CNN-based methods, which were designed for a limited number of diseases and often trained on small datasets [9]. Chen et al. (2022) presented an innovative approach for plant disease recognition by enhancing the YOLOv5 model [10]. They introduced novel techniques, such as Involution Bottleneck modules and SE modules, to improve the sensitivity of the model to channel features. Additionally, they modified the loss function to address its degeneration into 'Intersection over Union' [10]. Zhao et al. (2022) proposed the RIC-Net, a plant disease classification model that fused Inception and residual structures with an embedded attention mechanism. Their model achieved an impressive overall accuracy of 99.55% for identifying diseases in corn, potato, and tomato plants [11]. Abbas et al. (2021) tackled the issue of limited labeled data by leveraging Conditional Generative Adversarial Networks (C-GAN) to generate synthetic images of tomato plant leaves [12]. Khan et al. (2020) introduced a hierarchical framework for fruit disease classification, which combined deep features extraction, transfer learning, and multi-level fusion techniques [13].

Recently, novel approaches have emerged in the domain of transformers, which includes the integration of transformer models, such as Vision Transformers (ViTs), into the field of image-based disease identification [6]. Transformer-based models have gained significant traction across various applications, including natural language processing and computer vision [6]. Specifically, the Tokens-to-Token Vision Transformer (T2T-ViT) has demonstrated its effectiveness in enhancing the efficiency of ViTs and achieving superior performance in image classification tasks [6]. Furthermore, Convolutional Vision Transformers (CvT) amalgamate the advantages of convolutional neural networks (CNNs) and transformers, introducing convolutional layers into ViTs to enhance performance and efficiency [7]. These advancements in deep learning models offer promising solutions for addressing concerns related to food safety, particularly in the identification of plant diseases from images.

In recent literature, there has been a concerted effort towards developing innovative techniques to enhance feature extraction in image-based disease identification. However, it is crucial to acknowledge that as these techniques become more sophisticated, there is an associated increase in computational demands for accessing and analyzing the extracted features to make informed decisions. Our research primarily focuses on feature extraction using Vision Transformers (ViT) and aims to assess the impact of feature reduction on classification performance. We employ linear projection in various ways to explore its effects on performance parameters.

Additionally, we address the practical implementation of our proposed methodology by introducing a novel hardware design. This design is specifically tailored to facilitate the scanning of leaves for health determination. By incorporating hardware considerations, we aim to bridge the gap between theoretical advancements in feature extraction and their real-world applicability. Our holistic approach encompasses both algorithmic refinement and hardware optimization, contributing to a comprehensive and practical solution for automated plant disease identification.

## 3    Materials and Methods

### 3.1    Dataset

The Plant Village dataset [14], a significant contribution to plant disease detection, consists of 54,305 leaf images meticulously categorized into 38 distinct classes based on species and disease. Each image has a resolution of 256 px and is captured in RGB color space, with a uniform background for focused analysis. The dataset includes grayscale and background-removed variants, totaling 61,486 images after augmentation.

In our research, we focused on a subset featuring Apples, Corns, Peaches, Raspberries, Tomatoes, and Grapes. This subset comprises 3,306 images, showcasing both healthy and diseased states, with specific diseases such as Cedar Rust, Common rust, Black Measles, Bacterial spot and Yellow leaf curl virus.

## 3.2 Leaf diseases and their causes

Plant diseases have notable impacts on crops. Cedar Apple Rust, caused by Gymnosporangium juniperi-virginianae [15], affects apple trees, leading to rust-colored lesions on leaves due to fungal spores from cedar trees. Common Rust (Puccinia sorghi) [16] poses a threat to corn, manifesting as reddish-brown pustules on leaves. Grape Esca [17], affecting grapevines, results from fungal pathogens and environmental stress factors. Bacterial Spot, induced by Xanthomonas arboricola pv. Pruni [18], affects peach trees, causing dark lesions. Tomato plants are vulnerable to the Yellow Leaf Curl Virus (TYLCV) [19][20], transmitted by whiteflies, causing leaf curling and yellowing.

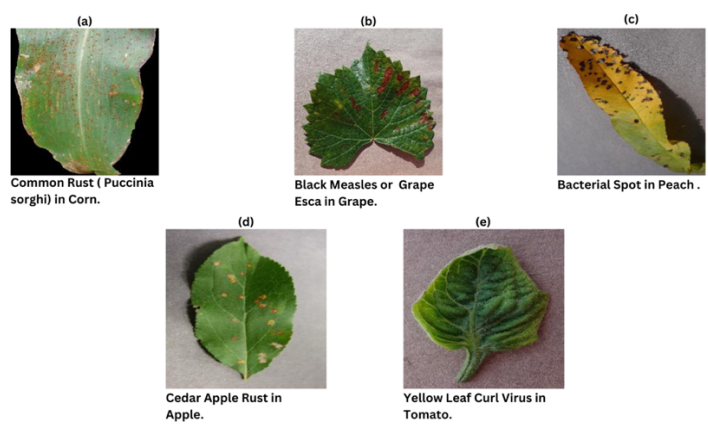

**Fig. 1.** (a) Common Rust; (b) Black Measles; (c) Bacterial Spot; (d) Cedar Apple Rust; (e) Yellow Leaf Curl Virus

## 3.3 Image Pre-Processing

In our research, we embarked on a meticulous image pre-processing workflow (see Fig. 2.) to enhance the robustness and effectiveness of subsequent image analysis methods.

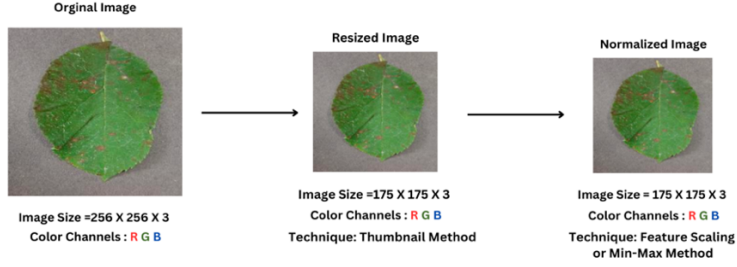

**Fig. 2.** Image Pre-Processing Workflow

The initial stage involved resizing the original image using the Thumbnail method [21]. First, the aspect ratio of the original image must be ascertained. This is done by dividing the height by the width (see Eqn. 1.).

$$Aspect\ Ratio\ (AR) = \frac{Original\ Height\ (H)}{Original\ Width\ (W)} \qquad (1)$$

One of the dimensions (either width or height) for the resized image is fixed based on the desired size of the thumbnail. Let's consider it as the target width $W_{target}$. Using the determined aspect ratio, the other dimension (height in this case) is computed (see Eqn. 2.).

$$New\ Height\ (H') = W_{target}\ X\ AR \qquad (2)$$

Often, during resizing, interpolation techniques are used to approximate pixel values for the resized image. One common method is bilinear interpolation. It determines the value of a new pixel based on a weighted average of the four pixels in the nearest 2x2 neighborhood. The weights are determined by the distance of the input pixel location to the center of each of the four pixels.

The image is then resized to $W_{target}\ X\ H'$ Dimensions while preserving the original aspect Ratio. Following this, we employed the Feature Scaling or Min-Max normalization technique. The Min-Max normalization for an individual pixel [22] can be described as in Eqn. 3.

$$p_{norm} = \frac{p_{orginal} - p_{min}}{p_{max} - p_{min}}\ X\ (new_{max} - new_{min}) + new_{min} \qquad (3)$$

Where $p_{norm}$ is the pixel value after normalization, $p_{orginal}$ is the original pixel value, $p_{max}$ and $p_{min}$ are the minimum and maximum pixel values in the original image, respectively. $new_{max}$ and $new_{min}$ are the desired bounds after normalization, typically set to 0 and 1 for image data.

Furthermore, to provide a comprehensive understanding of our pre-processing effects, we analyzed the pixel distribution across the Red, Green, and Blue (RGB) channels for both the original and normalized images. The juxtaposition of these pixel distributions reveals discernible shifts in frequency across various pixel values. Specifically, while the original image exhibits peaks indicative of dominant color shades, the normalized image showcases a more balanced spectrum. This balance indicates a more uniform pixel intensity distribution, enhancing the image's suitability for algorithms sensitive to variations in pixel values.

### 3.4    Feature Extraction using ViT (Vision Transformer)

The ViT Feature Extractor [23],[24] is a powerful neural network model designed for tasks like image classification, particularly in the context of diagnosing plant diseases. It reimagines the Transformer architecture, originally developed for natural language processing, and adapts it for image analysis.

The model begins by taking as input an image of a plant leaf infected with a disease. This image has a fixed size, specified by image size, and comprises three colour channels (R, G, B).

The input leaf image is divided into smaller, non-overlapping patches using a convolutional layer. This patch-based approach allows the model to focus on local information within the image, similar to how a plant disease expert might examine different parts of the leaf for disease symptoms.

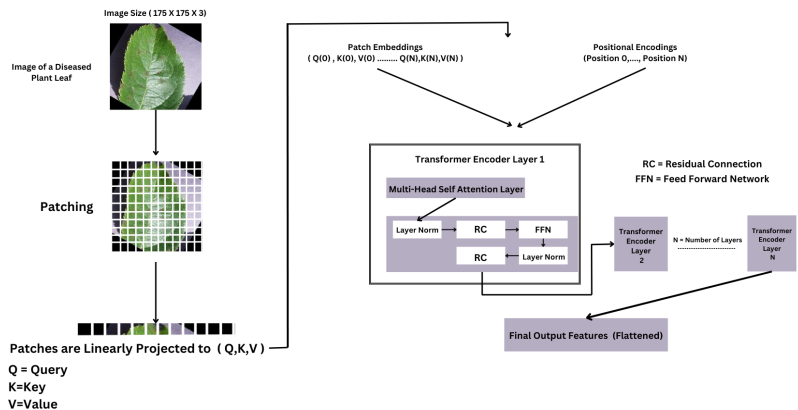

**Fig. 3.** Feature Extraction using ViT (Vision Transformer)

The preservation of spatial relationships between patches within an input image of a plant leaf infected with a disease is of utmost importance. To achieve this, positional embeddings are incorporated into the patch representations. These positional embeddings play a pivotal role in conveying critical spatial information about the precise location of each patch within the image. They essentially act as a reference system, aiding in the model's comprehension of the leaf's intricate structure.

In mathematical terms, we can express the fusion of positional embeddings as follows: P represents the positional embeddings linked to a specific patch, while X represents the patch's inherent features or characteristics.

Eqn. 4. illustrates the amalgamation of these two elements, creating a unified representation that combines the patch's inherent attributes with its spatial information in a cohesive manner.

$$Patch\_with\_Position = X + P \qquad (4)$$

This combined representation "$Patch\_with\_Position$", enriched with spatial information through positional embeddings, forms the foundation upon which the ViT Feature Extractor builds its understanding of the leaf's structure and, consequently, its capacity to identify and comprehend the location of potential issues within the leaf. In essence, this encoding mechanism ensures that the model possesses the requisite spatial awareness to effectively pinpoint areas within the leaf where anomalies or disease symptoms may be manifesting.

At the heart of the ViT (Vision Transformer) Feature Extractor (see Fig. 3.) lies a robust architectural framework consisting of multiple Transformer encoder layers. These layers serve as the bedrock upon which the model's ability to discern intricate patterns within plant leaf images is built. Within each of these encoder layers, a multi-head self-attention mechanism is applied. This mechanism is designed to mimic the way a botanist might inspect different parts of a leaf simultaneously, enabling the model to effectively concentrate on various regions within the image simultaneously.

The multi-head self-attention mechanism operates through the following mathematical formulation, where 'Q' represents query vectors, 'K' denotes key vectors, and 'V' stands for value vectors as in Eqn. 5.

$$Attention(Q, K, V) = softmax\left(\frac{Q\,K^T}{\sqrt[2]{d_k}}\right).V \qquad (5)$$

This attention mechanism allows the model to grasp how patches relate to one another, capturing both local and global dependencies within the image. However, to ensure the effective training of this deep architecture and mitigate the vanishing gradient problem, residual connections are introduced immediately after the self-attention step. These connections are instrumental in preserving the gradient flow, thereby facilitating the training process.

To further stabilize training and maintain consistent data distribution, layer normalization is applied to the output of the residual connection as in Eqn. 6. .

$$Layer\ Normalization(Residual\ Connection) \quad (6)$$

Following the application of self-attention and normalization, the data undergoes processing by a feedforward neural network (MLP). This network is analogous to how a botanist might analyze various characteristics of a leaf. It is proficient at capturing complex, non-linear patterns that are essential for understanding the leaf's condition. The structure of this feedforward network can be succinctly expressed as follows, where 'W1' and 'W2' denote weight matrices, 'b1' and 'b2' are bias vectors, and ReLU signifies the rectified linear unit activation function (as in Eqn. 7.).

$$FFN(Residual\_Connection) \\ = ReLU(W2 \cdot (W1 \cdot Residual\_Connection + b1) + b2) \quad (7)$$

The output of the Transformer encoder layers comprises a set of embeddings, one for each patch within the leaf image. These embeddings are enriched with both local and global information, effectively shaping feature representations of the leaf. They encode valuable information about the leaf's health, symptoms, and distinguishing characteristics.

To create a comprehensive feature representation of the entire leaf image, the embeddings are flattened into a single vector, encapsulating the combined knowledge extracted from all patches. This condensed vector serves as a holistic representation of the leaf's condition, making it possible for the disease's classification algorithms (in the next set of steps) such as Convolutional Neural Networks, to make informed decisions about the presence and distribution of disease symptoms and overall leaf health.

$$Flattened\_Embeddings = Flatten([Emb1, Emb2, \dots, EmbN]) \quad (8)$$

Here in Eqn. 8., $[Emb1, Emb2, \dots, EmbN]$ symbolizes the embeddings derived from each individual patch in the leaf image.

In essence, this architectural framework, with its attention mechanisms, residual connections, normalization, and MLPs, empowers the ViT Feature Extractor to comprehend the intricate details of plant leaves and extract valuable insights regarding their condition.

### 3.5    Feature Extraction using ViT (Vision Transformer) With Dimensionality Reduction Using Linear Projection

To the existing ViT Feature Extractor framework (see Fig. 4.), an additional linear projection layer has been integrated to effectively reduce the dimensionality of the feature vector. This modification addresses potential challenges associated with handling high-dimensional data, such as computational inefficiency and overfitting. By projecting the feature vector onto a lower-dimensional space, the model can focus on the most salient features of the image, leading to potentially better generalization during tasks like classification.

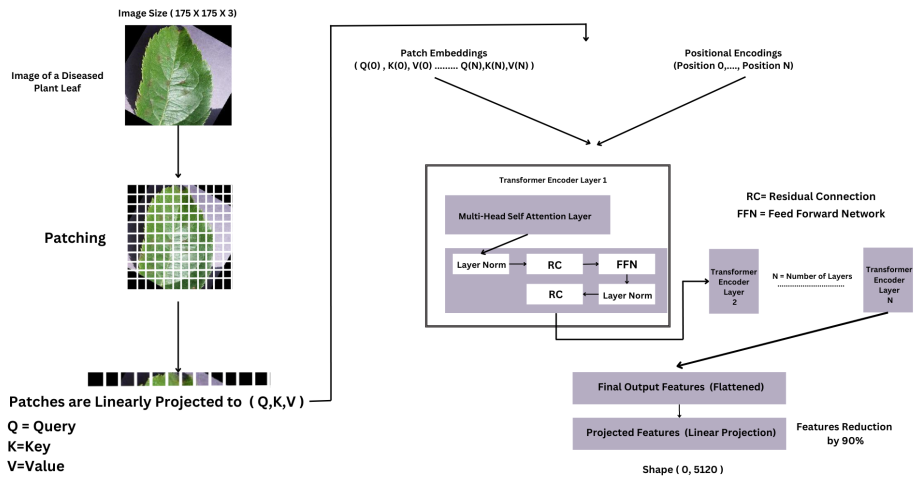

**Fig. 4.** Feature extraction using ViT (Vision Transformer) with dimensionality reduction layer

Mathematically, the linear projection operation can be visualized as a matrix multiplication that transforms the high-dimensional feature vector, $F$, into a lower-dimensional vector, $P$ (See Eqn. 9.).

$$P = F \ X \ W \quad (9)$$

Where, $F$ is the original high-dimensional feature vector derived from the flattened embeddings. $W$ represents the weight matrix associated with the dense layer used for projection. $P$ is the resulting projected (lower-dimensional) feature vector. The dense layer (or fully connected layer) used for this projection is defined as in Eqn. 10.

$$projected \ features = Dense \ (Number \ of \ Features) \ (Flattened\_Embeddings) \ (10)$$

Here, *Number Of Features* determines the dimensionality of the projected feature vector, *P*. The significance of this projection is that it allows the model to capture essential features from the high-dimensional space in a more compact representation.

### 3.6 Feature Extraction using ViT (Vision Transformer) With Block Wise Dimensionality Reduction Using Linear Projection

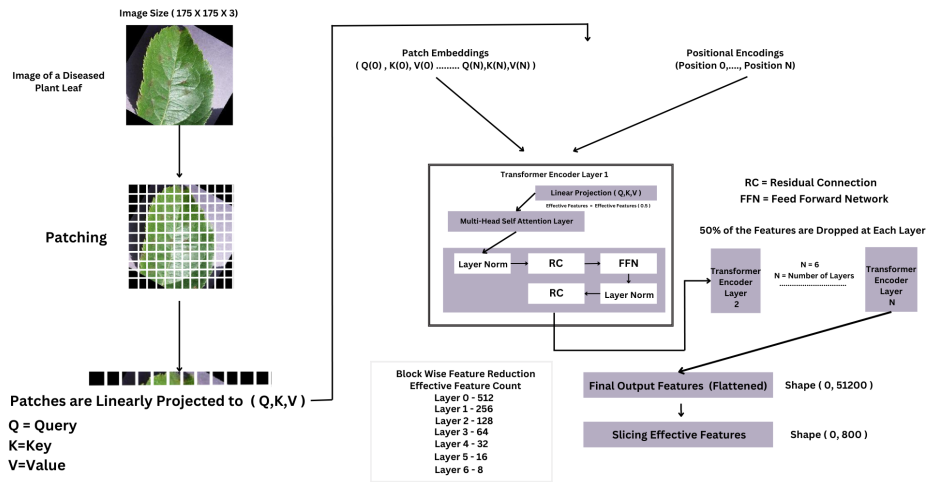

**Fig. 5.** Feature extraction using ViT (Vision Transformer) with Transformer Block Wise Dimensionality Reduction Layer

The enhanced Vision Transformer (ViT) introduces a linear projection layer within each transformer block (see Fig. 5.), a strategic modification aimed at tackling the challenges of high-dimensional data in deep learning. By compressing high-dimensional feature vectors into a lower-dimensional space at each stage, the model focuses on the most vital image features. This block-wise projection is crucial, as it allows for the distillation and capture of essential features, resulting in a compact yet informative representation. Such an approach is designed to improve computational efficiency and reduce the risk of overfitting, enhancing the model's performance in deep feature understanding and abstraction tasks like image classification. This innovation in the ViT feature extractor marks a significant advancement in handling and processing high-dimensional data in neural networks.

### 3.7 Hardware Implementation

In our hardware implementation, we utilize a novel design (See Fig. 6.) to scan leaves (both healthy and diseased) comprehensively. The setup includes a rotatable bed capable of a full 360-degree rotation for the leaf. An elevatable arm, equipped with a camera and adjustable up to 180 degrees, allows for variable-angle leaf capture. The entire system is powered by the Raspberry Pi Compute Module, with the camera utilizing Sony's IMX219 Sensor. Additionally, our design incorporates a rotatable LED strip to match ambient lighting conditions, enhancing visibility during scanning. Precise movements are achieved through Nema 17 Stepper motors, providing controlled and accurate positioning for effective leaf analysis.

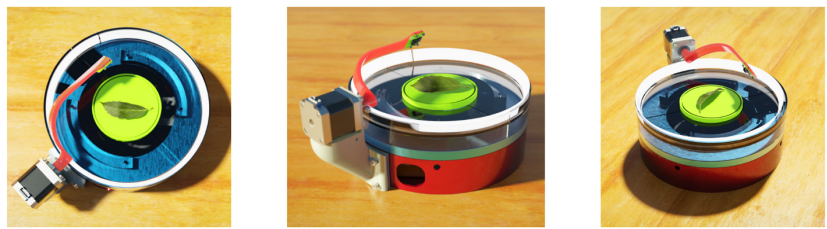

**Fig. 6.** Different views of the proposed novel hardware design

For automated plant disease identification, we have programmed the system to capture images of the leaf from various angles and orientations. Subsequently, these captured images undergo processing using the proposed framework, and features are extracted using ViT. These features are then employed in decision-making to determine whether the leaf is diseased or not, utilizing different CNN architectures as discussed in Section 4.

## 4    Model training

In an effort to predict diseases and identify healthy plants using feature data, two Convolutional Neural Network (CNN) architectures were implemented.

Two architectures (See Fig 7.) were designed for plant health prediction. "Architecture 1" starts with input reshaping, followed by two convolutional layers (32 and 64 filters) with max-pooling. It then flattens data, passes through a dense layer (128 neurons), a dropout layer (rate 0.5), and ends with an output layer (softmax activation).

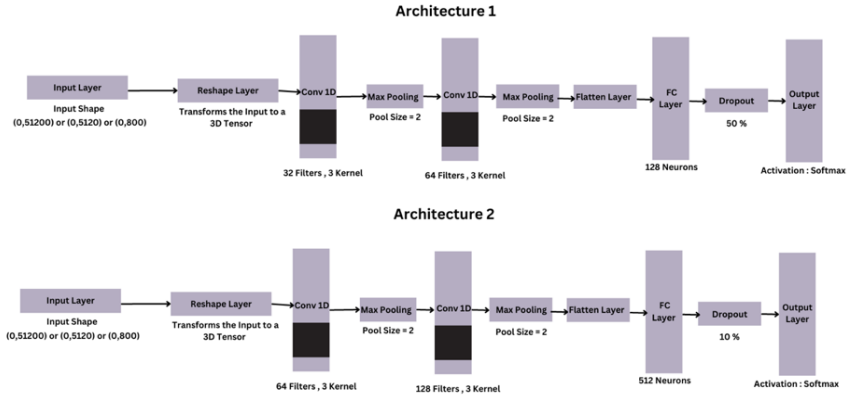

**Fig. 7.** Illustrations for "Architecture 1" and "Architecture 2" CNN models.

"Architecture 2" has a similar structure but with differences in filter count (64 and 128) and neurons in the dense layer (512). Dropout rate is 0.1. Both models use Adam optimizer (learning rate = 0.001), sparse_categorical_crossentropy loss, and ModelCheckpoint for optimal weights. Training involves a batch size of 32, a maximum of 50 epochs, and Early Stopping (patience=25).

"Architecture 1" and "Architecture 2" were trained using three feature sets: higher-order features for intricate pattern understanding, linearly projected dimensionally reduced features for generalization, and block-wise linear projection features for a balanced approach. This comprehensive training strategy ensures a holistic view, combining detailed analysis, generalization, and iterative refinement to optimize overall performance.

## 5    Results and Discussion

The evaluation of the models in this study involved a comprehensive assessment using various performance metrics to gauge their classification capabilities. The metrics employed for this evaluation include Precision (P), Recall (R), and F1-score (F1) for each individual class. These metrics provide a detailed perspective on the models' adeptness in classifying instances for specific categories.

Table 1. Evaluation Metrics for Architecture 1 Without Linear Projection.

| Model | Class | Precision | Recall | F1-Score | Support |
|---|---|---|---|---|---|
| | Apple – Cedar Apple Rust | 0.90 | 0.87 | 0.88 | 30 |
| | Apple-Healthy | 0.90 | 0.90 | 0.90 | 30 |
| | Corn – Common Rust | 1.00 | 1.00 | 1.00 | 30 |

| Model | Class | Precision | Recall | F1-Score | Support |
|---|---|---|---|---|---|
| Architecture 1 (Without Linear Projection) | Corn – Healthy | 0.97 | 1.00 | 0.98 | 30 |
| | Grape -Black Measles | 0.97 | 0.97 | 0.97 | 30 |
| | Grape-healthy | 0.93 | 0.93 | 0.93 | 30 |
| | Peach-Bacterial Spot | 0.96 | 0.87 | 0.91 | 30 |
| | Peach-healthy | 0.94 | 0.97 | 0.95 | 31 |
| | Raspberry- Healthy | 0.84 | 0.87 | 0.85 | 31 |
| | Tomato - YLCW | 1.00 | 1.00 | 1.00 | 30 |
| | Tomato - Healthy | 0.90 | 0.93 | 0.92 | 30 |
| Micro-Averaged Metrics | | 0.94 | 0.94 | 0.94 | |
| Macro-Averaged Metrics | | 0.94 | 0.94 | 0.94 | |
| Hamming Loss | | | | | 0.06 |

Table 2. Evaluation Metrics for Architecture 2 Without Linear Projection.

| Model | Class | Precision | Recall | F1-Score | Support |
|---|---|---|---|---|---|
| Architecture 2 (Without Linear Projection) | Apple – Cedar Apple Rust | 0.966 | 0.933 | 0.949 | 30 |
| | Apple-Healthy | 0.906 | 0.967 | 0.935 | 30 |
| | Corn – Common Rust | 1.00 | 1.00 | 1.00 | 30 |
| | Corn – Healthy | 0.968 | 1.00 | 0.984 | 30 |
| | Grape -Black Measles | 0.967 | 0.967 | 0.967 | 30 |
| | Grape-healthy | 0.964 | 0.900 | 0.931 | 30 |
| | Peach-Bacterial Spot | 0.90 | 0.90 | 0.90 | 30 |
| | Peach-healthy | 0.966 | 0.903 | 0.933 | 31 |
| | Raspberry- Healthy | 0.933 | 0.933 | 0.933 | 31 |
| | Tomato - YLCW | 0.938 | 1.00 | 0.968 | 30 |
| | Tomato - Healthy | 0.90 | 0.90 | 0.90 | 30 |
| Micro-Averaged Metrics | | 0.95 | 0.95 | 0.95 | |
| Macro-Averaged Metrics | | 0.946 | 0.946 | 0.945 | |
| Hamming Loss | | | | | 0.054 |

Table 3. Evaluation Metrics for Architecture 1 With Linear Projection.

| Model | Class | Precision | Recall | F1-Score | Support |
|---|---|---|---|---|---|
| Architecture 1 (With Linear Projection) | Apple – Cedar Apple Rust | 0.931 | 0.900 | 0.915 | 30 |
| | Apple-Healthy | 0.897 | 0.867 | 0.881 | 30 |
| | Corn – Common Rust | 1.00 | 1.00 | 1.00 | 30 |
| | Corn – Healthy | 1.00 | 1.00 | 1.00 | 30 |
| | Grape -Black Measles | 0.853 | 0.967 | 0.906 | 30 |
| | Grape-healthy | 0.806 | 0.967 | 0.879 | 30 |
| | Peach-Bacterial Spot | 0.778 | 0.933 | 0.848 | 30 |
| | Peach-healthy | 0.967 | 0.935 | 0.951 | 31 |
| | Raspberry- Healthy | 0.864 | 0.633 | 0.731 | 31 |
| | Tomato - YLCW | 0.897 | 0.867 | 0.881 | 30 |
| | Tomato - Healthy | 1.00 | 0.867 | 0.929 | 30 |
| Micro-Averaged Metrics | | 0.90 | 0.90 | 0.90 | |
| Macro-Averaged Metrics | | 0.908 | 0.903 | 0.90 | |
| Hamming Loss | | | | | 0.097 |

Table 4. Evaluation Metrics for Architecture 2 With Linear Projection.

| Model | Class | Precision | Recall | F1-Score | Support |
|---|---|---|---|---|---|
| Architecture 2 | Apple – Cedar Apple Rust | 0.929 | 0.867 | 0.897 | 30 |
| | Apple-Healthy | 0.966 | 0.933 | 0.949 | 30 |
| | Corn – Common Rust | 1.00 | 1.00 | 1.00 | 30 |
| | Corn – Healthy | 1.00 | 1.00 | 1.00 | 30 |
| | Grape -Black Measles | 0.906 | 0.967 | 0.935 | 30 |
| | Grape-healthy | 0.967 | 0.967 | 0.967 | 30 |
| (With Linear Projection) | Peach-Bacterial Spot | 0.750 | 0.900 | 0.818 | 30 |
| | Peach-healthy | 0.935 | 0.935 | 0.935 | 31 |
| | Raspberry- Healthy | 0.897 | 0.867 | 0.881 | 31 |
| | Tomato - YLCW | 0.867 | 0.867 | 0.867 | 30 |
| | Tomato - Healthy | 0.962 | 0.833 | 0.893 | 30 |
| Micro-Averaged Metrics | | 0.92 | 0.92 | 0.92 | |
| Macro-Averaged Metrics | | 0.925 | 0.921 | 0.922 | |
| Hamming Loss | | | | | 0.079 |

Table 5. Evaluation Metrics for Architecture 1 With Block Wise Linear Projection.

| Model | Class | Precision | Recall | F1-Score | Support |
|---|---|---|---|---|---|
| | Apple – Cedar Apple Rust | 0.846 | 0.733 | 0.786 | 30 |
| | Apple-Healthy | 0.875 | 0.933 | 0.903 | 30 |
| | Corn – Common Rust | 1.00 | 1.000 | 1.00 | 30 |
| | Corn – Healthy | 1.00 | 1.00 | 1.00 | 30 |
| Architecture 1 | Grape -Black Measles | 0.875 | 0.933 | 0.903 | 30 |
| (With Block | Grape-healthy | 0.903 | 0.933 | 0.918 | 30 |
| Wise | Peach-Bacterial Spot | 0.802 | 1.000 | 0.938 | 30 |
| Linear Projection) | Peach-healthy | 0.794 | 0.900 | 0.844 | 31 |
| | Raspberry- Healthy | 0.897 | 0.839 | 0.867 | 30 |
| | Tomato - YLCW | 0.923 | 0.800 | 0.857 | 30 |
| | Tomato - Healthy | 0.812 | 0.867 | 0.839 | 30 |
| Micro-Averaged Metrics | | 0.89 | 0.89 | 0.89 | |
| Macro-Averaged Metrics | | 0.890 | 0.888 | 0.887 | |
| Hamming Loss | | | | | 0.112 |

Table 6. Evaluation Metrics for Architecture 2 With Block Wise Linear Projection.

| Model | Class | Precision | Recall | F1-Score | Support |
|---|---|---|---|---|---|
| | Apple – Cedar Apple Rust | 0.933 | 0.933 | 0.933 | 30 |
| | Apple-Healthy | 0.867 | 0.867 | 0.867 | 30 |
| | Corn – Common Rust | 1.000 | 1.000 | 1.000 | 30 |
| | Corn – Healthy | 1.000 | 1.000 | 1.000 | 30 |
| Architecture 2 | Grape -Black Measles | 0.909 | 1.000 | 0.952 | 30 |
| (With Block | Grape-healthy | 0.879 | 0.967 | 0.921 | 30 |
| Wise Linear | Peach-Bacterial Spot | 0.839 | 0.867 | 0.852 | 30 |
| Projection) | Peach-healthy | 0.903 | 0.903 | 0.903 | 31 |
| | Raspberry- Healthy | 0.889 | 0.800 | 0.842 | 31 |
| | Tomato - YLCW | 0.923 | 0.800 | 0.857 | 30 |
| | Tomato - Healthy | 0.900 | 0.900 | 0.900 | 30 |
| Micro-Averaged Metrics | | 0.91 | 0.91 | 0.91 | |
| Macro-Averaged Metrics | | 0.913 | 0.912 | 0.912 | |
| Hamming Loss | | | | | 0.088 |

In an extensive evaluation of two distinct neural network architectures, performance variances were noted across three different settings, particularly in relation to the implementation of linear projection.

In the initial setting without linear projection (Table 1. And Table 2.), Architecture 1 demonstrated robust performance with micro and macro-averaged metrics at 0.94, and a low Hamming loss of 0.06. It achieved perfect F1-scores for classes like Corn – Common Rust and Tomato - YLCW. Architecture 2, also without linear projection, showed marginally superior performance with micro-averaged metrics of 0.95 and macro-averaged metrics of 0.946, coupled with a lower Hamming loss of 0.054.

Upon introducing linear projection (Table 3. And Table 4.), Architecture 1 saw a decline in performance metrics, with both micro and macro-averaged values dropping to 0.90 and an increased Hamming loss of 0.097. This decline was uneven across classes, with some like Grape - Black Measles improving, while others like Raspberry - Healthy declined. Meanwhile, Architecture 2 maintained a relatively high performance with linear projection, recording micro and macro-averaged metrics of 0.92 and 0.925, respectively, and a Hamming loss of 0.079. This performance, though reduced from its non-linear projection state, still surpassed that of Architecture 1 under similar conditions.

In a third setting (Table 5. And Table 6.), Architecture 1 achieved micro and macro-averaged metrics of 0.88 and 0.883, respectively, with a Hamming loss of 0.112. Architecture 2 outperformed with a precision, recall, and F1-score of 0.91 each, a macro-averaged metric of 0.913, and a reduced Hamming loss of 0.088.

Overall, while both architectures showed commendable performances, the absence of linear projection in Architecture 2 led to the most effective results. The introduction of linear projection slightly attenuated the performance in both architectures, yet Architecture 2 consistently outperformed its counterpart in all scenarios.

# 6    Conclusion

Our research delved into utilizing the ViT (Vision Transformer) Feature Extractor for plant disease diagnostics, employing image segmentation into patches and incorporating positional embeddings. We integrated dimensionality reduction to enhance computational efficiency and model generalization. Two CNN architectures, "Architecture 1" and "Architecture 2," were trained and evaluated on both original and linearly projected features. The study revealed the general superiority of "Architecture 2," especially in scenarios without linear projection. However, the introduction of linear projection led to a performance decrease for both architectures, although "Architecture 2" maintained its advantage.

Noteworthy findings include the consistent high accuracy of "Architecture 2" in classifying diseases like "Corn – Common Rust." Variances in performance across different classes, such as "Raspberry- Healthy," highlight the importance of tailored architectural choices. The study conclusively demonstrates the efficacy of ViT in plant disease diagnostics, with "Architecture 2" emerging as the more robust CNN model. It is essential to balance these findings against computational resource considerations and specific task requirements for practical implementations, ensuring an optimal blend of accuracy and efficiency in real-world applications.

Additionally, our novel hardware design is specifically tailored for scanning diseased leaves in an omnidirectional fashion. The hardware implementation utilizes a Raspberry Pi Compute Module to address low memory configurations, ensuring practicality and affordability. This innovative hardware solution enhances the overall feasibility and accessibility of the proposed automated disease identification system.

**Code Availability:**
Yes, Custom Code.

**CRediT author statement:**
Abhishek Sebastian - Conceptualization, Methodology, Software, Writing – Original Draft, Project administration.
Annis Fathima A - Supervision, Writing – Review & Editing.
Pragna R - Conceptualization, Methodology, Writing – Original Draft.
Madhan Kumar S - Formal analysis, Investigation, Software and Data Curation.
Yaswanth Kannan G - Software, Visualization and Formal analysis.
Vinay Murali – Resources , Methodology , Conceptualization and Visualization.